\documentclass[conference]{IEEEtran}
\IEEEoverridecommandlockouts
\usepackage{cite}
\usepackage{amsmath,amssymb,amsfonts}
\usepackage{algorithmic}
\usepackage{graphicx}
\usepackage{textcomp}
\usepackage{subfigure}
\usepackage{xcolor}
\def\BibTeX{{\rm B\kern-.05em{\sc i\kern-.025em b}\kern-.08em
    T\kern-.1667em\lower.7ex\hbox{E}\kern-.125emX}}
\begin{document}

\title{Corner Case Data Description and Detection}

\author{

\IEEEauthorblockN{1\textsuperscript{st} Tinghui Ouyang}
\IEEEauthorblockA{\textit{Artificial Intelligence Research Center (AIRC)},\\
\textit{National Institute of Advanced Industrial Science and Technology (AIST)},
Tokyo, Japan\\
ouyang.tinghui@aist.go.jp}

\and
\IEEEauthorblockN{2\textsuperscript{nd} Vicent Sanz Marco}
\IEEEauthorblockA{\textit{AIRC},
\textit{AIST},
Tokyo, Japan \\
vicent.sanzmarco@aist.go.jp}
\and
\IEEEauthorblockN{3\textsuperscript{rd} Yoshinao Isobe}
\IEEEauthorblockA{\textit{Cyber Physical Security Research Center(CPSRC)},
\textit{AIST},
Tsukuba, Japan\\
y-isobe@aist.go.jp}
\and
\IEEEauthorblockN{4\textsuperscript{th} Hideki Asoh}
\IEEEauthorblockA{\textit{AIRC},
\textit{AIST},
Tokyo, Japan \\
h.asoh@aist.go.jp}
\and
\IEEEauthorblockN{5\textsuperscript{th} Yutaka Oiwa}
\IEEEauthorblockA{\textit{Cyber Physical Security Research Center(CPSRC)},
\textit{AIST},
Tsukuba, Japan \\
y.oiwa@aist.go.jp}
\and

\IEEEauthorblockN{6\textsuperscript{th} Yoshiki Seo}
\IEEEauthorblockA{\textit{AIRC},\textit{AIST},Tokyo, Japan \\
y.seo@aist.go.jp}

}

\maketitle

\begin{abstract}
As the major factors affecting the safety of deep learning models, corner cases and related detection are crucial in AI quality assurance for constructing safety- and security-critical systems. The generic corner case researches involve two interesting topics. One is to enhance DL models’ robustness to corner case data via the adjustment on parameters/structure. The other is to generate new corner cases for model retraining and improvement. However, the complex architecture and the huge amount of parameters make the robust adjustment of DL models not easy, meanwhile it is not possible to generate all real-world corner cases for DL training.  Therefore, this paper proposes a simple and novel approach aiming at corner case data detection via a specific metric. This metric is developed on surprise adequacy (SA) which has advantages on capture data behaviors. Furthermore, targeting at characteristics of corner case data, three modifications on distanced-based SA are developed for classification applications in this paper. Consequently, through the experiment analysis on MNIST data and industrial data, the feasibility and usefulness of the proposed method on corner case data detection are verified.
\end{abstract}


\section{Introduction}

As the fast development of machine learning technologies in the last decade, especially deep learning (DL) based techniques, this makes Artificial Intelligent (AI) a widely-known term in both industries and our daily life. However, this fast development also makes us concerning AI applications, e.g the safety of DL-based systems \cite{b1}. As more complex architectures of deep neural networks are applied to obtain high accuracy in applications, the safety  becomes more critical than ever before, especially in life- and property-related fields, like medical diagnosis \cite{b2}, malware detection \cite{b3}, autonomous driving \cite{b4} and so on. For instances, autonomous driving is being developed in many big companies, like Tesla, Ford, Waymo/Google, but several serious accidents have happened in real-world due to DLs’ safety issues \cite{b5,b6}. These accidents imminently grab scholars' attention around world on studies about finding factors affecting the safety in DL systems testing and taking measurements to eliminate them.

In both traditional software and DL-based systems, incorrect or unexpected corner case behaviors are always regarded to play an important part affecting systems’ safety \cite{b7}. For example, in autonomous driving, many reported real-world collision cases are related to those rare or previously unseen corner cases. In \cite{b8}, a collision between a Tesla car and a trailer was reported since the Tesla’s DL system failed to deal with the corner cases of “white color against a brightly lit sky” and the “high ride height”. Another collision was reported on a Google self-driving car crashed to a bus since its DL system made wrong decision on rare conditions (corner cases) \cite{b9}. Besides the given two examples, many other wrong decisions made by AI systems are more or less related to different corner cases. Therefore, to detect and fix those potential flaws or undesired corner case behaviors, namely corner case detection, is surely crucial for system designers to construct safety- and security-critical DL systems, and it should also be a necessary part of systematical AI testing.

Currently, many scholars are studying on corner cases and related AI testing techniques \cite{b10,b11}. Firstly, from the perspective of concepts, corner cases are namely those erroneous behaviors in the DL-based software, which are analogous to bugs in traditional software \cite{b12}. Different with traditional software, corner cases of DL systems pay more attention on data distribution instead of system architectures. Hence, traditional software usually aims at patch fixing after bug detection, DL-based systems need to consider these detected erroneous corner case data in the AI model retraining process for improving its robustness on structure and parameters. Secondly, researches about corner cases study mainly include the following two parts. One kind of study is to test a given DL model’s robustness and stability when facing with corner case data. For example, some adversarial attack strategies, such as in \cite{b13,b14}, are applied on the original testing data to generate adversarial samples which can reflect the feature of corner cases to some extent, then DL model’s accuracy and robustness are tested. The DeepFool proposed in \cite{b15} can also realize the generation of adversarial corner cases and robustness testing. The DeepMutation proposed in \cite{b16} aimed at mutating DL models’ structure and parameters directly, and testing the modified model’s performance on the original testing data. Moreover, through this mutation and testing, some undetected corner cases can be further detected. For example, if a data point is found erroneous after a small model mutation, it is possible to be regarded as a potential corner case data due to its high risk of causing wrong decision. However, considering current industrial DL systems always having thousands of neurons and millions of parameters, it is extremely challenging to detect corner cases via introducing small mutations or perturbations on models. The other kind of study is to generate corner case data for retraining and improving DL systems’ performance. For instance, the DeepMutation technique also proposed some mutation operators on data, and generated some corner cases for DL model testing. In \cite{b17}, biased corner cases were also generated from MNIST dataset based on the metamorphic testing technique. Moreover, the DeepTest method proposed in \cite{b7} aimed at the autonomous driving scenario, and generated lots of corner cases by leveraging image transformation techniques to change the driving conditions, like rain, fog and lighting.  Furthermore, the DeepXplore method \cite{b19} proposed an idea of neuron coverage, and made use of it to iteratively learn erroneous corner cases. There were thousands of incorrect corner case behaviors generated in this paper. However, even though companies like Tesla and Google have developed many effective techniques in corner case study, it is still hardly possible to generate and consider all kinds of corner cases in the training process. The only feasible way would be studying a way for detecting corner cases as possible in AI testing.

Concerning the mentioned techniques and problems on corner case study, this paper proposes to develop a novel corner case data detection method by using modified \emph{distance-based surprise adequacy}  (DSA). The idea of surprise adequacy (SA) was initially proposed by Kim et al \cite{b20}, which can be used as a test adequacy tool for DL system testing. SA’s initial property is to describe the surprise of testing data with respect to the training data, namely to describe difference/similarity between testing and training data.  In \cite{b20}, two kinds of SA are developed, one based on the probability density distribution, the other based on distance similarity, and they were furthermore verified useful in industrial applications \cite{b21,b22}. While, through the review on SA-related literature, two useful points can be obtained. One is that DSA is specifically effective for classification applications. The other is that SA was verified useful to capture data’s behaviors in DL testing, namely models’ neuron activation behaviors instead of plain data distribution, for example, in \cite{b21} SA was validated to correlate with the correctness answers in NLP study. With consideration of these two points, this paper is inspired to leverage DSA for corner case detection in classification problems, since SA’s capability on data description can be used not only on normal data but also on incorrect/erroneous data (corner case data). Moreover, to enhance DSA’s ability on capturing behaviors of corner case data, this paper also proposes three kinds of modification on the DSA definition. Furthermore, based on the proposed DSA and the corner case detection method, experiments based on a benchmark classification data (MNIST) \cite{b23} and industrial application data are implemented and studied in this paper. Consequently, the novelty and contributions of this paper can be summarized as follows:
\begin{itemize}
\item SA is applied for data description, especially for corner case data description. Instead of describing data via the general plain data distribution directly, SA adapts neurons activation behaviors with respect to a DL model to describe data’s characteristics, which can capture profound data behaviors responding to DL models in quality testing;
\item Three kinds of modification on DSA definitions are developed. First, the capability of DSA on data behaviors description is inherited. Then, according to the idea that erroneous corner cases are tightly related to the classifier boundary, so modifications on the DSA definition are proposed;
\item Based on DSA, a novel corner case data detection method is proposed. Different from most of corner case study, the proposed method can be utilized as a tool in recognition of corner case data. In this way, it is possible to detect any corner case data with no need to generate/learn all possible corner case data;
\item Experiments on MNIST data and industrial data validate the feasibility of using DSA to describe corner case data behaviors, and that it is useful to use DSA on corner case data detection. Moreover, the proposed DSA3 achieve a relatively better performance than others based on experiment results analysis.
\end{itemize}

The rest of the paper is organized as follows. Section II briefly describes the idea of surprise adequacy and its related calculation process. Section III gives out three kinds of modification based on corner case definition and their characteristics. Section IV implements experiments on the benchmark MNIST data and an industrial classification data, and analyze the performance of using DSA on corner case data detection. Finally, Section V concludes the contributions of this paper, and think about the future work.
\section{Surprise Adequacy}

To evaluate the quality of testing data in AI assurance, a good way is to study neurons behavior in terms of a given deep learning (DL) model. Generally, the more the diversity of neuron behaviors, the better the quality of testing data. There are various methods to describe neuron behaviors in DL models. For example, neuron coverage was proposed in \cite{b19}, and its activation status was taken as the behavior of a testing data point responding to the DL model. Similarly, values and signs of neuron outputs can both be used to describe neurons’ activation behaviors in \cite{b24}.

While, compared with those metrics reflecting independent behaviors of testing and training sets, in \cite{b20} an interesting idea was proposed to describe the difference between the testing set’s behaviors and that of the whole training set. To realize this idea, the activation status of neurons in DL models are still firstly denoted. Assuming a set of inputs  $X=\{x_1,x_2,...\}$ and a trained DL model M consisted of a set of neurons $\boldsymbol{N}=\{n_1,n_2,...\}$. For a given testing data $x\in X$ and an ordered (sub)set of neurons $N\subseteq\boldsymbol{N}$, the activation behavior (namely \emph{activation trace}) of $x$ on $N$ is expressed by
the vector of activation values and it denoted as
\begin{equation}
  \alpha_N (x)=[a_1 (x),a_2 (x),..., a_{|N|} (x)]^T\label{e1}
\end{equation}
where
each element $a_n(x)$ corresponds to the activation value of $x$ with respect to an individual neuron $n$ in $N$.
Hence, the set of activation traces for $X$ is denoted as
$A_N(X) = \{\alpha_N(x)~|~x\in X\}$.

Then, $A_N(Tr)$ is calculated based on  the training dataset $Tr$, which records neurons’ activation behaviors on all samples in $Tr$. Subsequently, the activation behavior of testing data $Te$ is also obtained as $A_N(Te)$. 
Finally, combining $A_N(Tr)$ and $A_N(Te)$, surprise adequacy (SA) was defined to describe the relative novelty of testing inputs with respect to the training data. It is actually denoted as the quantitative similarity measure between $A_N(Tr)$ and $A_N(Te)$ in the following form.
\begin{equation}
SA=SimilarilryMeasure(A_N (Te),A_N (Tr))\label{e2}
\end{equation}

  In \cite{b20}, there are two kinds of similarity measurement proposed to formalize SA. One is the likelihood-based SA (LSA), and the other one is distance-based SA (DSA). These two SAs are validated feasible to capture the relative surprise of testing data in DL systems, and describe testing data’s behaviors, which implies SAs are useful to evaluate the quality of testing data.
\section{Corner cases detection based on SA}
\subsection{Denotation of corner cases}

\begin{figure}[htbp]
\centerline{\includegraphics[scale=0.3]{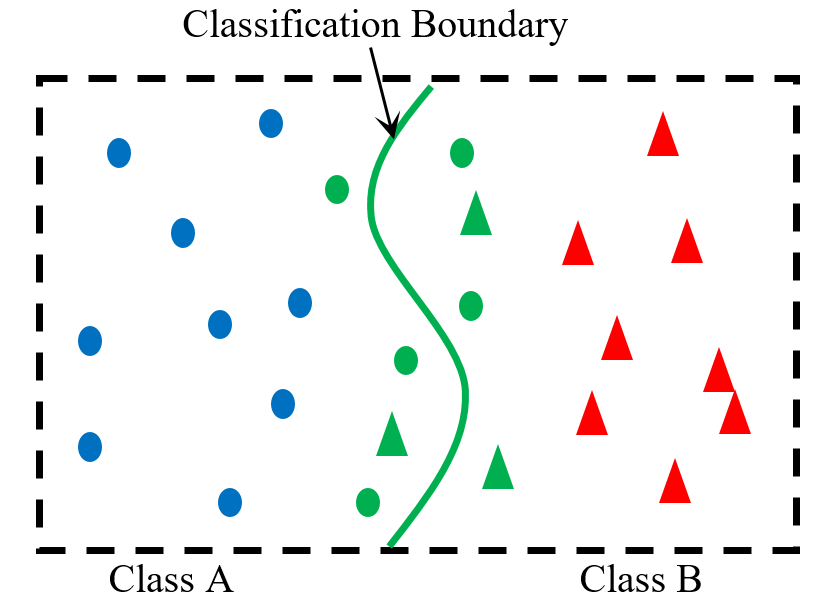}}
\caption{Diagram of corner case. Two class of data are colored as blue and red. Data of corner case is colored as green, which includes incorrectly classified data as well as data close to classification boundary which is sensitive to cause unexpected recognition. }
\label{f1}
\end{figure}

As well as the execution of traditional software \cite{b25}, a dangerous condition in AI system testing is processing data of corner cases which generally cause incorrect and unexpected behaviors. For example, when DL-based autonomous driving system processes corner cases of rainy weather or strong reflection, an incorrect decision may be made to lead to crash bringing the loss of life and property. Therefore, detecting corner-case samples is important in AI testing. According to the above description of corner case, we can define the corner case set as the following form
\begin{equation}
Corner\ case\ set:\{x| \ DL(x+pertubation)\neq label(x)\}\label{e3}
\end{equation}
where, $x$ is denoted as a sample in corner case; its true label is denoted as $label(x)$; $DL(*)$ is the output class based on a given DL model. Through this definition, we see when a small perturbation is added into a corner-case data $x$, where $perturbation$ is a small value such that $0<{|perturbation|}\leq \varepsilon $ for a constant $\varepsilon$ , the class recognized by DL system will be different with its true label. In this way, a corner case set can include data samples with both incorrect and unexpected behaviors, e.g. boundary adversarial data and incorrectly classified data (outliers), as shown in Fig. \ref{f1}.

\subsection{Distance-based SA and its modification}

Considering SA can capture the behavior of testing data, therefore we may think about the possibility of using SA to describe behaviors of data in corner case. Here, we mainly consider the distance-based SA (DSA), since it describes the relation of testing data respect to the classification boundary. According to the DSA definition in \cite{b20}, firstly assuming a DL model M consisting of a set of neurons $N$, it is learned based on a training dataset $Tr$ with $C$ classes. According to the denotation of SA, the activation traces of Tr can be calculated as $A_N(Tr)$ based on the trained M. Then, for a new testing input $x$ in the class $c_x\in C$, its nearest neighbor in $c_x$ is denoted as $x_a$. Finally, the surprise of $x$ respect to class $c_x$ can be calculated as the distance between their activation trace $\alpha_N(*)$, as the following form
%
\begin{equation}
  x_a=\underset{x_i\in X~\mbox{\small s.t.}~c_{x_i}=c_x}{\operatorname{argmin}}\Vert \alpha_N (x)-\alpha_N (x_i)\Vert       \label{e5}
\end{equation}
\begin{equation}
  dist_a= 
  \Vert \alpha_N (x)-\alpha_N (x_a)\Vert\label{e4}
\end{equation}
where, $\|*\|$ is the Euclidean distance. Subsequently, taking $x_a$ as the reference point and finding its nearest neighbor $x_b$ in a class different with $c_x$, then the surprise between $x_a$ and $x_b$ is also calculated 
\begin{equation}
x_b=\underset{x_i\in X~\mbox{\small s.t.}~c_{x_i}\neq c_x}{\operatorname{argmin}}\|\alpha_N (x_a )-\alpha_N (x_i)\|    \label{e6}     
\end{equation}
\begin{equation}
dist_b=\|\alpha_N (x_a )-\alpha_N (x_b)\| \label{e7}
\end{equation}
%
Combining the above definitions, the SA of testing data $x$ to the training data is consequently defined as the ratio between $dist_a$ and $dist_b$, as below
\begin{equation}
DSA(x)=\frac{dist_a}{dist_b}\label{e8}
\end{equation}

\begin{figure}[htbp]
\centerline{\includegraphics[scale=0.4]{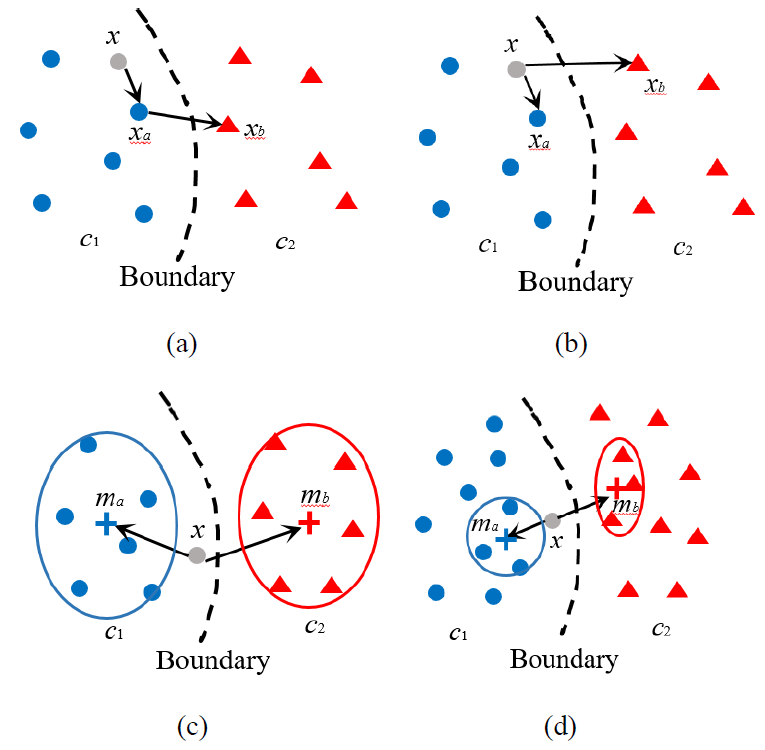}}
\caption{Diagram of four types of DSA. (a) the original DSA. (b) modification 1. (c) modification 2. (d) modification 3.}
\label{f2}
\end{figure}

According this definition, it is seen if the numerator is larger and the denominator is smaller, the value of DSA will be larger, implying the testing data $x$ is surprise to data of class $c_x$ in the training set, as shown in Fig. \ref{f2}(a). Therefore, we can see that DSA shows a way of describing data’s surprise to training data by distance, and it is also useful to describe data’s activation behaviors respect to the given DL model.
While, the above definition of DSA is simple, it has some drawbacks to deal with independent data points in extreme cases. Moreover, to evaluate its capability on detecting data of corner case, several additional modifications are also proposed in this paper to formalize DSA for corner case detection.

(1) Modification 1: novelty calculation via testing data itself

In the original DSA definition, we see that the denominator is actually to calculate the surprise of $x_a$ (the nearest neighbor of the testing data $x$ in the same class) respect to data of other different classes. Therefore, DSA can be regarded as the comparison between $x$’s novelty in its belonging class and its class novelty to other classes. This definition may be useful to describe the surprise of a testing data to the whole training data. However, to evaluate if a data sample belongs to corner case, its own novelty respect to all classes seems more important. Hence, we modify the original DSA by just adjusting the definition of $dist_b$ as the following form
\begin{equation}
x_b=\underset{x_i\in X~\mbox{\small s.t.}~c_{x_i}\neq c_x}{\operatorname{argmin}} \|\alpha_N (x)-\alpha_N (x_i)\| \label{e9}
\end{equation}
\begin{equation}
dist_b=\|\alpha_N (x)-\alpha_N (x_b)\|\label{e10}
\end{equation}

The calculation of DSA keeps unchanged. This modification makes DSA consider surprises of testing data $x$ to all classes independently, which may be helpful to reflect the behaviors of data in corner case, as shown in Fig. \ref{f2}(b).

(2) Modification 2: novelty calculation via global data descriptors

Compared with the original DSA, the first modification considers surprise of testing data $x$ to data of all classes. While, these two DSA definitions may have a common shortage on processing pair-wise rare data points, especially on describing behaviors of corner case. For example, assuming the testing data $x$ is the corner-case data, there happens to exist a neighbor $x_a$ very close to $x$, namely $x\approx x_a$, since the above DSA are calculated based on the distance of point-to-point, this may lead DSA to be a low value  due to $dist_a\rightarrow0$.  In this case, the novelty of $x$ respect to training data is regarded as small based on SA definition. However, this result causes the inconsistence of behavior novelty for data in corner case. Furthermore, this result will be meaningless for corner case study as well. Therefore, a possible modification on the basis of the first modification is to select a more general class descriptor for surprise calculation. In \cite{b26}, the class centers were commonly used as feasible global descriptors. Hence, we can replace the nearest neighbor with the class center in surprise calculation, and the new equations can be written as below
\begin{equation}
dist_a=\|\alpha_N (x)-\alpha_N (m_a )\|    \label{e11}
\end{equation}
\begin{equation}
dist_b=\|\alpha_N (x)-\alpha_N (m_b )\|  \label{e12}
\end{equation}
where
$m_a$ is the center of the class 
which $x$ belongs to;
$m_b$ is the center of a different class $c_{b} \ ( c_{b} \neq c_x)$
which
the closest point $x_b$ to $x$ belongs to.
For each $s \in \{a,b\}$, the center point $m_s$ is
calculated by
  
\begin{equation}
  m_s = \frac{1}{|X_s|} \sum_{x_i\in X_s}{x_i}, ~\mbox{where}~ X_s=\{x_i\in X|\ c_{x_i}=c_s\}.
\label{e13}
\end{equation}
Then, the newly modified DSA can be calculated as $dist_a/dist_b$. Its diagram is shown in Fig. \ref{f2}(c).

(3) Modification 3: novelty calculation via local data descriptors

In the above modification of DSA, the characteristic of a class is simply described by class center. However, corner-case data usually have some obvious characteristics of closing to boundary or outlier location. Class centers are good global descriptors of classes, but not effective to capture local characteristics, e.g. in reflecting the characteristics of corner case data. In this paper, to modify DSA for corner case data detection, we propose to replace the global descriptors with local descriptors, e.g. the center of
a nearest neighborhood
shown in Fig. \ref{f2}(d). The DSA of testing data $x$ can be still calculated via \eqref{e11}, \eqref{e12} and \eqref{e8},
while the calculation of the center
$m_{s,\delta}$ for each $s \in \{a,b\}$ should be modified as 
\begin{equation}
m_{s,\delta}=\frac{1}{|X_{s,\delta}|} \sum_{x_i\in X_{s,\delta}} {x_i}     
\label{e14}
\end{equation}
where $X_{s,\delta}$ ($\subseteq X_s$) represents the neighborhood set of inputs
included in the hypersphere whose center is $x_s$ and radious is $\delta$,
namely several nearest neighbors of $x_s$.
Here, two methods determining the neighborhood set are provided. One kind of neighborhood is denoted by a given size of Eculidean round
(i.e. the same radius $\delta$ for all inputs is used), 
as follows 
\begin{equation}
X_{s,\delta} = \{x_i\in X_s~|~\|x_i-x_s\| < \delta\}   
\end{equation}
where, $x_a$ ($s=a$) and $x_b$ ($s=b$) are inputs explained in the previous Modification 2 and
$X_s$ is the set defined in \eqref{e13}.  
An alternative definition of the nearest neighborhood is determined by $k$-nearest points \cite{b27}.
In this case, the radious $\delta$ is given for each $x_s$ such that 
$X_s$ includes only $k$ nearest neighbors of $x_s$ (i.e. $|X_{s,\delta}|=k$) 
and therefore
\begin{equation}
  X_{s,\delta}=\{x_{i_1},x_{i_2},..., x_{i_k}\}  
  \label{e15m}
\end{equation}
where, $x_{i_j} \in X_s$ is the $j_{th}$ nearest point to $x_s$.  

\subsection{Corner case data coverage} 
Based on the above definitions of DSA, including the original one and modifications, the analysis on studying DSA’s behaviors responding to corner case data can be implemented. To quantitatively evaluate the capability of DSA in corner case data detection, here we propose a metric as corner-case data coverage as the following form
\begin{equation}
cov(v_{th} )=\frac{card(\{d|\ d\in CD,\ DSA(d)>v_{th}\})}{|CD|}*100\%
\label{e16}
\end{equation}
where, $v_{th}$ is a given threshold; $CD$ represents the dataset of corner case data; $|CD|$ is the cardinality; $\{d\}$ represents the set consisting of all detected corner case data which have DSA values larger than the given threshold. While, considering not all of data in corner case are detectable, here we only take the detectable corner-case data into account, namely those data wrongly recognized by DL. In this way, the proposed corner case data coverage is actually to evaluate the percentage of erroneous behaviors of DL systems.
\section{Experiments and Evaluation} 
\subsection{Dataset and the DL system description} 
To study behaviors of the proposed DSA on corner case data detection, in this paper we take the MNIST data \cite{b23} as a reference in experiments and analysis. MNIST is a widely-used dataset in machine learning research, which contains ten classes of images, and dividing as 60,000/10,000 training/testing samples. For recognizing different numbers in MNIST, we adopt a common-used five layer Convolutional Neural Network (CNN), including convolutional, max-pooling, dropout and full-connection layers. 
On the other hand, experiments in this paper are implemented on a machine equipped with Intel i7-9750 CPU, 32GB RAM, NVIDIA GeForce RTX 2060, running in Windows 10 Pro 64-bit OS. Code of this paper is released \cite{b30}, which is developed from SADL \cite{b28} on Keras, and contains the proposed three kinds of modification of DSA are realized.
\subsection{Qualitative analysis on corner case data detection} 
\begin{figure}[htbp]
\centerline{\includegraphics[scale=0.3]{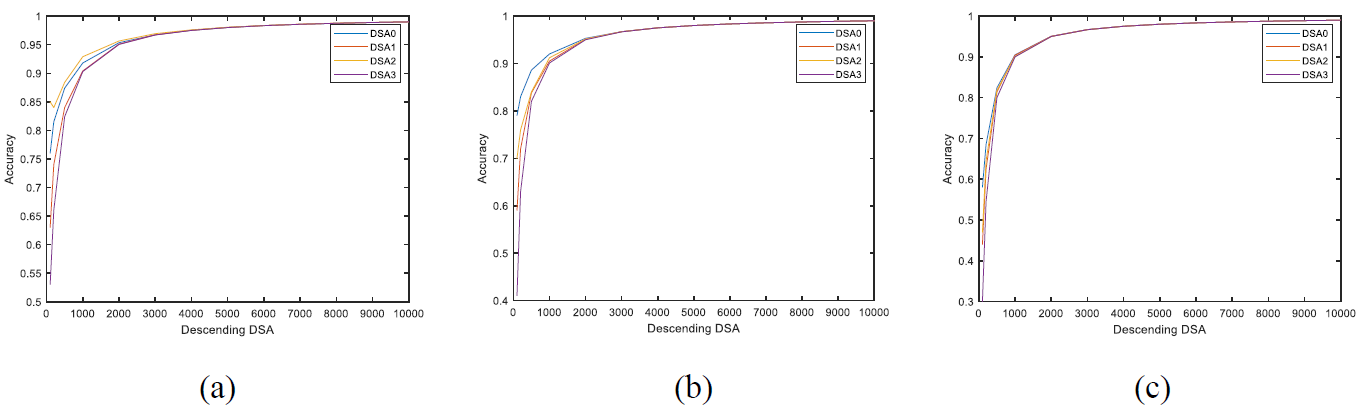}}

\caption{Testing accuracy variation as the descending DSA. (a) Layer1. (b) Layer2. (c) Layer3.}
\label{f3}
\end{figure}

Firstly, through training the given CNN architecture above on MNIST data, we can achieve $99.17\%$ accuracy on the original testing set. Then, based on the proposed DSA definition and the trained DL model, we can calculate the values of DSA on each testing sample, and subsequently study DSA’s behaviors responding to corner case data detection.
By sorting all testing samples as the DSA descending trend, we can first study the influence of DSA variation on DL model’s performance. In order to do comparative analysis, four kinds of DSA are calculated on outputs of three selected layers, such as the convolutional layer 1 (Layer1), convolutional layer 2 (Layer2) and the final full-connection layer (Layer3) of the given CNN. Results are shown in Fig. \ref{f3}.

\begin{figure}[htbp]
\centerline{\includegraphics[scale=0.35]{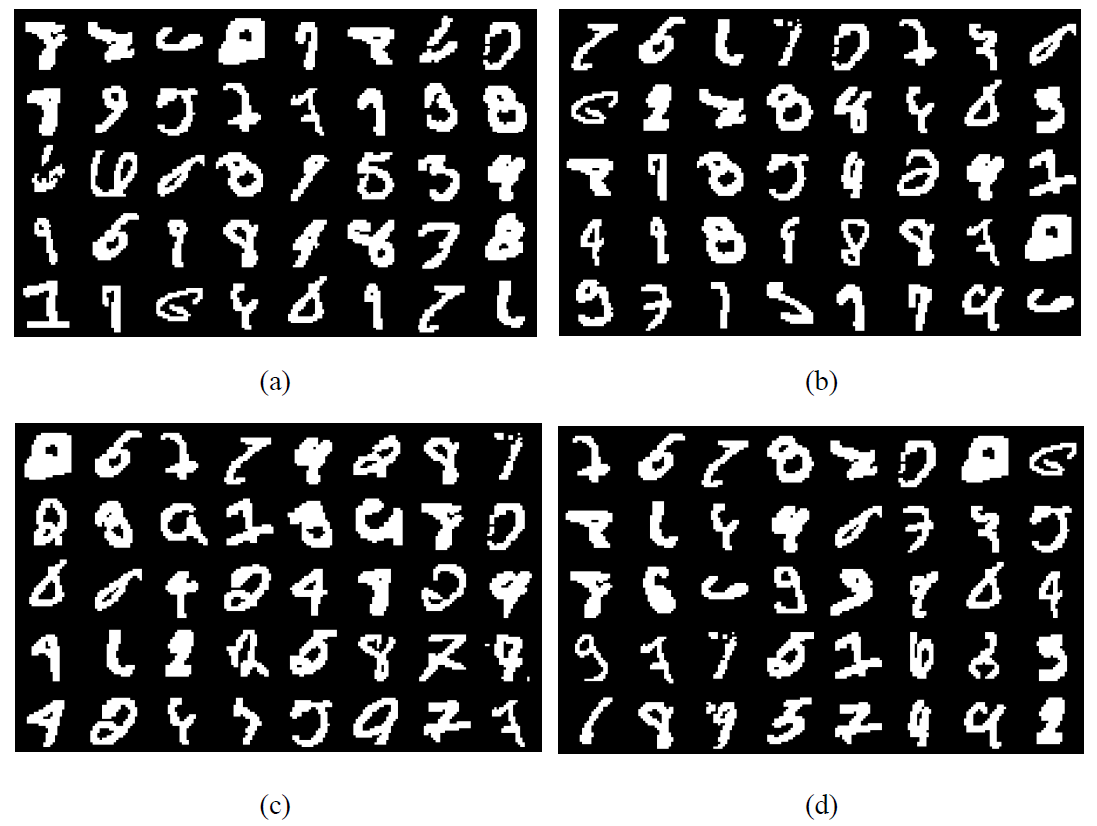}}

\caption{Images with the largest DSA values. (a) DSA0. (b) DSA1. (c) DSA2. (d) DSA3.}
\label{f4}

\end{figure}

Fig. \ref{f3} shows the curves of accuracy vs. DSA variation. By assuming the first $k_n$ 
samples are taken from the sorted dataset for testing, then an accuracy value is calculated.
As the value of $k_n$ 
ascends from a small value to 10,000, the curve of accuracy can be plotted. Here, the beginning point is $k_n=100$, where we can see the accuracy is low due to a certain percentage of erroneous data (corner case data). This phenomenon can be found in all curves during the beginning period where testing samples have large DSA values. This implies that large values of DSA can capture corner case data to some extent. Then, the ending point is $k_n=10,000$ meaning the whole testing dataset, all curves reach the final testing accuracy $99.17\%$. Moreover, to vividly analyze this foundation, we can further draw out images of these samples with the largest DSA values, as shown in Fig. \ref{f4}.

In Fig. \ref{f4}, images are drawn separately based on four DSA definitions, such as DSA0- DSA3 representing the original DSA and other three modifications respectively, and DSA3 adopts the second definition in \eqref{e15m} which is more convenient in computation (e.g. $k$=20). From these images, we can see there are indeed many corner case data, which are caused by various factors, like unusual writing, thick chirography, large slant, structure disequilibrium and so on. Some of the samples can be unrecognized by not only AI but also human beings, however they can have large value of DSA. Therefore, it illustrates that DSA might be useful to detect corner case data.
\subsection{Quantitative analysis of DSA behaviors} 
\begin{figure}[htbp]
\centerline{\includegraphics[scale=0.3]{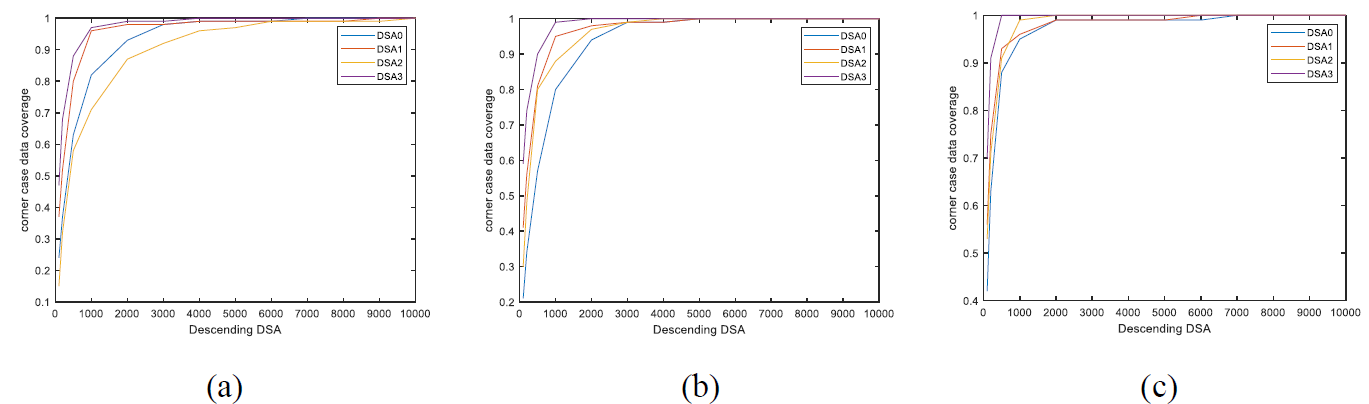}}

\caption{Corner case data coverage of testing data sorted as the descending DSA. (a) Layer1. (b) Layer2. (c) Layer3.}
\label{f5}
\end{figure}

While, besides the conclusion that DSA can capture activation behaviors of data from Fig. \ref{f3}, there discovers the other interesting phenomenon, namely DSA3 has relatively lower accuracy in all these three layers, especially at the beginning period of curves. This implies DSA3 can capture more wrongly recognized images to lower the accuracy when DSA values are selected as a large value, which is also to say DSA3 has a better capability on corner case data detection. To verify this conjecture and to quantitatively study performance of different DSAs on corner case data detection, the proposed corner-case data coverage is firstly calculated. Results of corner-case data coverage based on four DSAs are shown in Fig. \ref{f5}.

In Fig. \ref{f5}, the testing data is also sorted by the descending direction of DSA values. For a given DSA threshold $v_{th}$, the number of data satisfying $DSA(x)>v_{th}$ is counted. Meanwhile, under this condition the number of wrongly classified data is also counted for the calculation of corner-case data coverage. It is seen from these results, that DSA3 has the largest corner-case data coverage on the analysis of all three given layers.
On the other hand, we can ignore the selection of threshold as a concert value, and utilize the measurement of AUC-ROC \cite{b29} on corner case data detection to evaluate different DSAs’ performance. By assuming to utilize DSA values directly for distinguish normal data and corner case data, the AUC value of this binary classification process can be calculated. AUC-ROC values of four DSAs in three given layers are presented in Table \ref{t1}.
\begin{table}
\centering
\caption{AUC-ROC of corner case data detection based on DSA}\label{t1}

\begin{tabular}{|c|c|c|c|}
\hline
 &Layer1&Layer2&Layer3\\
\hline
DSA0&0.9422&0.9418&0.9759\\
\hline
DSA1&0.9657&0.9722&0.9833\\
\hline
DSA2&0.9114&0.9662&0.9875\\
\hline
DSA3&0.9814&0.9884&0.9966\\
\hline

\end{tabular}
\end{table}

Based on the result in Table \ref{t1}, three conclusions can be obtained. First, for the constructed CNN model on MNIST, it indeed achieves good performances by using DSA to guide corner case data detection since all AUC values are larger than 0.9. Second, it is found that the deeper the layer is, the better DSA performs on corner case data detection, seeing $AUC(Layer3)>AUC(Layer2)>AUC(Layer1)$. Third, DSA3 can have a relatively better performance when compared to the other DSA definitions. Therefore, based on these results, we can consider to choose the proposed DSA3 for corner case data detection in the further study.
\subsection{Adversarial testing analysis} 
\begin{figure}[htbp]
\centerline{\includegraphics[scale=0.4]{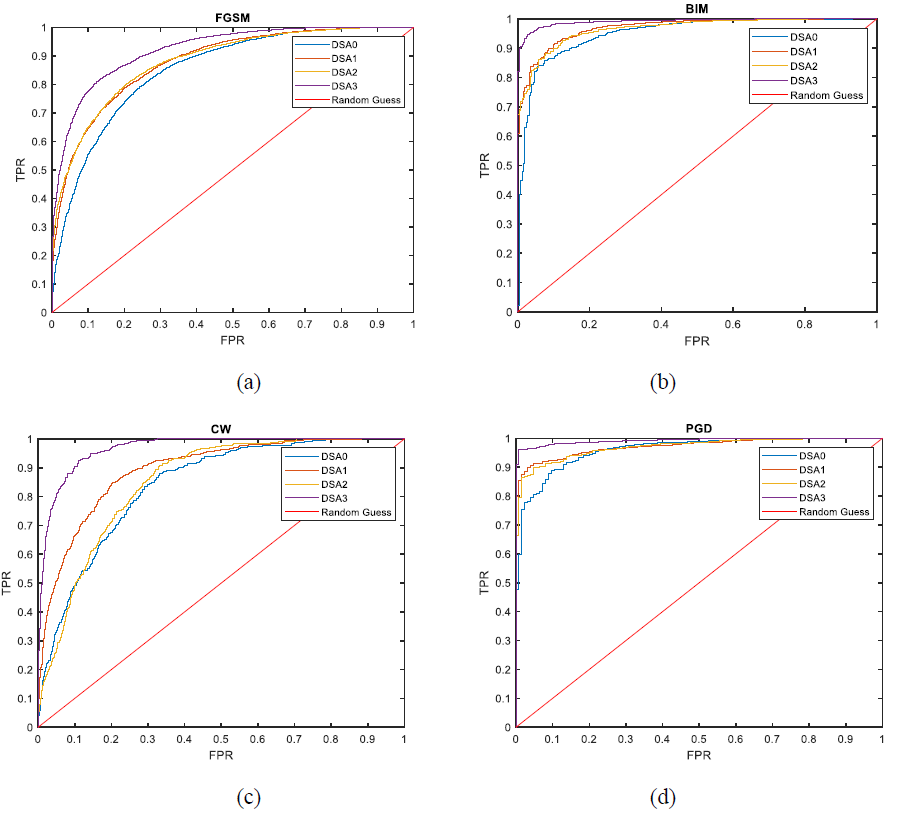}}

\caption{ROC of using DSAs on corner case data detection on four adversarial testing data. (a) FGSM. (b) BIM. (c) CW. (d) PGD.}
\label{f6}
\end{figure}

To further analyze the performance of DSA, as an important AI testing technique adversarial testing can also be applied here. Generally, based on the original testing data, small perturbations imperceptible to humans are added to generate adversarial samples which will lead the DL model to make wrong behaviors (outputs). There are many different attack strategies applied to generate adversarial testing set in literature. In this paper, we take four widely-studied adversarial attacks to generate new testing set on MNIST, such as Fast Gradient Sign Method (FGSM) \cite{b14}, Basic Iterative Method (BIM) \cite{b31}, Carlini $\&$ Wagner $(C\&W)$ \cite{b32}, and Projected Gradient Descent (PGD) \cite{b33}. Then, based on these four adversarial testing datasets, the proposed DSAs are applied to detect the corner case data, namely misclassified data. Similar as the above analysis, we do not care about the exact selection of DSA threshold, and just utilize the AUC-ROC to evaluate the performance of using DSA on corner case data detection. The results of ROC curves of four DSAs on different adversarial testing sets are depicted in Fig. \ref{f6}.

Fig. \ref{f6} shows the ROC curves on corner case data detection, where X-axis represents the value of FPR, and Y-axis represents the value of TPR respectively. In details \cite{b34}, TPR means true positive rate, also called as sensitivity or recall, which can be used to reflect the test's ability on correctly detecting corner case data here; FPR means false positive ratio, also known as false alarm ratio, which is the probability of falsely rejecting the null hypothesis for a particular test. Their calculation are presented as below
\begin{equation}
TPR=\frac{TP}{TP+FN} \label{e17}
\end{equation}
\begin{equation}
FPR=\frac{FP}{FP+TN}\label{e18}
\end{equation}
where, TP (True Positive), FP (False Positive), FN (False Negative), TN (True Negative) are four events in confusion matrix of corner case data detection. Moreover, based on these ROC curves, we can quantitatively calculate the values of AUC as a KPI metric, shown in Table \ref{t2}

\begin{table}
\centering
\caption{AUC-ROC of DSA-based corner case data detection on adversarial data testing.}\label{t2}

\begin{tabular}{|c|c|c|c|c|}
\hline
 &FGSM&BIM&CW&PGD\\
\hline	
DSA0	&0.8520	&0.9510	&0.8390	&0.9593\\
\hline
DSA1	&0.8801	&0.9702	&0.8931	&0.9722\\
\hline
DSA2	&0.8812	&0.9659	&0.8510	&0.9696\\
\hline
DSA3	&\textbf{0.9226}	&\textbf{0.9915}	&\textbf{0.9666}	&\textbf{0.9921}\\
\hline

\end{tabular}
\end{table}

According to the results of Table \ref{t2}, the same foundation is obtained that DSA3 performs relatively better than the other DSA definitions on all these four adversarial testing sets. Therefore, we can make use of DSA3 in the following study related to corner case data and AI quality assurance.

\subsection{Real-world data analysis}
\begin{figure}[htbp]
\centerline{\includegraphics[scale=0.3]{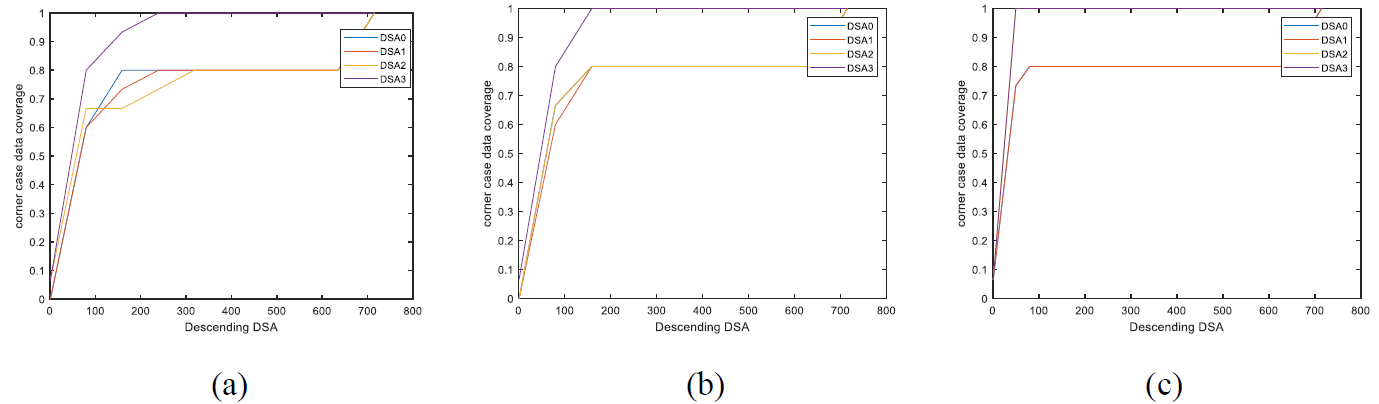}}

\caption{Corner case data coverage variance as the descending DSA. (a) Layer1. (b) Layer2. (c) Layer3.}
\label{f7}
\end{figure}

To further study the feasibility and generalization ability of the proposed method, we also propose to apply the proposed DSA method to detect corner case data in real-world application. Here, the metal casting product image data is taken for case study, which is applied for industrial quality inspection in real world \cite{b35}.  It has totally 6,633 training data images, and 715 testing images. The objective based on this data is an identical binary classification problem which mainly aims at automatically recognizing defective metal casting products and normal ones. Therefore, a general CNN model consisting of 2 convolutional layers and 2 full-connection layers are constructed for this issue. After modeling and training, the final testing accuracy can reach $97.90\%$ on this casting data. Then, according the proposed method on DSA calculation and corner case data detection, we can also implement related experiments, such as calculating four kinds of DSA values on given model layers, e.g. two convolutional layers and the output layer are considered here and also named as Layer1, Layer2, Layer3, corner case data coverage analysis, and performance analysis about using DSA on corner case data detection. Results of related experiments are presented in Fig. \ref{f7}.

In Fig. \ref{f7}, the corner case data coverage is calculated and plotted as the descending values of four given DSA. We can see the proposed DSA3 has relatively larger corner case data coverage, namely implying better capability on corner case capture.
\begin{figure}[htbp]
\centerline{\includegraphics[scale=0.3]{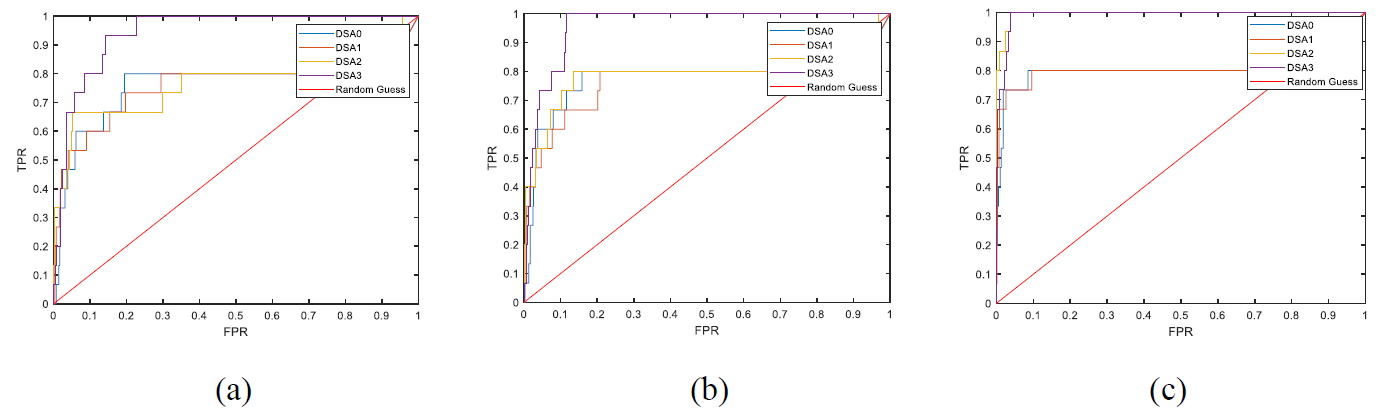}}

\caption{ROC curves of using DSAs on corner case data detection. (a) Layer1. (b) Layer2. (c) Layer3.}
\label{f8}
\end{figure}

Fig. \ref{f8} shows the ROC curves of using DSA on corner case data detection. It is also found that DSA3 has larger values of TPR under the same value of FPR than other DSAs, the similar phenomenon is shown in all three given layers analysis. For the further quantitative analysis, based on the results of ROC curves, we also calculate the values of AUC-ROC, as presented in Table \ref{t4}.
\begin{table}[htbp]
\caption{AUC-ROC of DSA-based corner case data detection on casting defect data}\label{t4}
\centering
\begin{tabular}{|c|c|c|c|}
\hline
	&Layer1	&Layer2	&Layer3\\
\hline
DSA0	&0.7565	&0.7717	&0.7990\\
\hline
DSA1	&0.7520	&0.7612	&0.8002\\
\hline
DSA2	&0.7584	&0.7815	&0.9952\\
\hline
DSA3	&0.9437	&0.9587	&0.9908\\
\hline
\end{tabular}
\end{table}

From results of Table \ref{t4}, we find the similar conclusion as that obtained on MNIST data that the proposed DSA modifications can improve the capability of the original DSA on describing corner case data’s behaviors, meanwhile the DSA3 performs relatively better than others. Moreover, as the layer becomes deeper, the calculated DSA has better performance on corner case data detection, seeing that $AUC(Layer3)>AUC(Layer2)>AUC(Layer1)$.
\section{Conclusions} 
In this paper, a novel method based on DSA has been proposed for corner case data detection which would be useful in AI quality assurance, such as safety analysis. While, to improve the capability of the original DSA on capturing behaviors of corner case data, three DSA modifications have been also developed with the consideration of boundary features, and validated useful in corner case data detection. Then, based on the experiments on MNIST which has been regarded as the benchmark classification application, it is seen that large values of DSA indeed capture many images with abnormal characteristics with respect to standard numbers, implying the DSA’s capability on capturing corner cases. Subsequently, through the quantitative analysis on erroneous corner case data detection, the modified DSAs have been found to have relatively better performance than the original one. Meanwhile, the proposed DSA3 performs the best on MNIST to some extent. Moreover, four adversarial testing datasets of MNIST have been also generated to further verify the feasibility of using DSA on corner case data detection. On the other hand, the industrial data based on casting defect application has also been studied in experiments, and its results further verify the general feasibility of using the proposed idea on classification problems.

While, besides the obtained conclusions above, more issues would be still meaningfully developed in the future work. First, exploring other possible metrics for corner case data detection, e.g. uncertainty consideration, since the  proposed DSAs in this paper cannot always perform perfectly. Second, studying the usage of detected corner case data, some possible studies like AI robustness analysis, stability and dependability in AI quality assurance, may be able to make use of these corner case data.  

\section*{Acknowledgment}

This research is based on results obtained from a project 'JPNP20006', commissioned by the New Energy and Industrial Technology Development Organization (NEDO)



\begin{thebibliography}{00}

\bibitem{b1} M. Borg, C. Englund, and B. Duran, “Traceability and Deep Learning-Safety-critical Systems with Traces Ending in Deep Neural Networks,” In Proc. of the Grand Challenges of Traceability: The Next Ten Years, 2017, pp. 48-49.
\bibitem{b2} S. Ellahham, N. Ellahham, and M. C. E. Simsekler, “Application of artificial intelligence in the health care safety context: opportunities and challenges,” American Journal of Medical Quality, vol. 35, no. 4, 2020, pp. 341-348. 
\bibitem{b3} Z. Yuan, Y. Lu, Z. Wang, and Y. Xue, “Droid-sec: deep learning in android malware detection,” In Proceedings of the 2014 ACM conference on SIGCOMM, pp. 371-372, August 2014.
\bibitem{b4} S. Liu, L. Liu, J. Tang, B. Yu, Y. Wang, and W. Shi, “Edge computing for autonomous driving: Opportunities and challenges,” Proceedings of the IEEE, vol. 107, no. 8, 2019, pp. 1697-1716. 
\bibitem{b5} D. Rice, “The Driverless Car and the Legal System: Hopes and Fears as the Courts, Regulatory Agencies, Waymo, Tesla, and Uber Deal with this Exciting and Terrifying New Technology,” Journal of Strategic Innovation and Sustainability, vol. 14, no. 1, 2019.
\bibitem{b6} S. Grigorescu, B. Trasnea, T. Cocias, and G. Macesanu, “A survey of deep learning techniques for autonomous driving,” Journal of Field Robotics, vol. 37, no. 3, 2020, pp. 362-386.
\bibitem{b7} Y. Tian, K. Pei, S. Jana, and B. Ray, “Deeptest: Automated testing of deep-neural-network-driven autonomous cars,” In Proceedings of the 40th international conference on software engineering, pp. 303-314, May 2018. 
\bibitem{b8} tesla-accident 2016. “Understanding the fatal Tesla accident on Autopilot and the NHTSA probe,” https://electrek.co/2016/07/01/
\bibitem{b9} google-accident 2016. “A Google self-driving car caused a crash for the first time,” http://www.theverge.com/2016/2/29/11134344/google-selfdriving-car-crash-report. 
\bibitem{b10} W. Wu, H. Xu, S. Zhong, M. R. Lyu, and I. King, “Deep validation: Toward detecting real-world corner cases for deep neural networks,” In 2019 49th Annual IEEE/IFIP International Conference on Dependable Systems and Networks (DSN), IEEE, pp. 125-137, June 2019.
\bibitem{b11} J. A. Bolte, A. Bar, D. Lipinski, and T. Fingscheidt, “Towards corner case detection for autonomous driving,” In 2019 IEEE Intelligent Vehicles Symposium (IV), IEEE, pp. 438-445, June 2019. 
\bibitem{b12} C. J. Clemente, F. Jaafar, and Y. Malik, “Is predicting software security bugs using deep learning better than the traditional machine learning algorithms? ,” In 2018 IEEE International Conference on Software Quality, Reliability and Security (QRS), IEEE, pp. 95-102, July 2018.
\bibitem{b13} A. Nguyen, J. Yosinski, and J. Clune, “Deep neural networks are easily fooled: High confidence predictions for unrecognizable images,” In Proceedings of the IEEE conference on computer vision and pattern recognition, pp. 427-436, 2015.
\bibitem{b14} I. J. Goodfellow, J. Shlens, and C. Szegedy, “Explaining and Harnessing Adversarial Examples,” In Proceedings of the 3rd International Conference on Learning Representations, 2015. http://arxiv.org/abs/1412.6572
\bibitem{b15} S. M. Moosavi-Dezfooli, A. Fawzi, and P. Frossard, “Deepfool: a simple and accurate method to fool deep neural networks,” In Proceedings of the IEEE conference on computer vision and pattern recognition, pp. 2574-2582, 2016.
\bibitem{b16} L. Ma, F. Zhang, J. Sun, M. Xue, B. Li, F. Juefei-Xu, and Y. Wang, “Deepmutation: Mutation testing of deep learning systems,” In 2018 IEEE 29th International Symposium on Software Reliability Engineering (ISSRE), IEEE, pp. 100-111, October 2018. 
\bibitem{b17} S. Nakajima, and T. Y. Chen, “Generating biased dataset for metamorphic testing of machine learning programs,” In IFIP International Conference on Testing Software and Systems Springer, Cham, pp. 56-64, October 2019.

\bibitem{b19} K. Pei, Y. Cao, J. Yang, and S. Jana, “Deepxplore: Automated whitebox testing of deep learning systems,” In proceedings of the 26th Symposium on Operating Systems Principles, pp. 1-18, October 2017.
\bibitem{b20} J. Kim, R. Feldt, and S. Yoo, “Guiding deep learning system testing using surprise adequacy,” In 2019 IEEE/ACM 41st International Conference on Software Engineering (ICSE), IEEE, pp. 1039-1049, May 2019.
\bibitem{b21} S. Kim, and S. Yoo, “Evaluating Surprise Adequacy for Question Answering,” In Proceedings of the IEEE/ACM 42nd International Conference on Software Engineering Workshops, pp. 197-202, June 2020.
\bibitem{b22} J. Kim, J. Ju, R. Feldt, and S. Yoo, “Reducing DNN Labelling Cost using Surprise Adequacy: An Industrial Case Study for Autonomous Driving,” 2020. arXiv preprint arXiv:2006.00894.
\bibitem{b23} Y. LeCun, L. Bottou, Y. Bengio, and P. Haffner. "Gradient-based learning applied to document recognition." Proceedings of the IEEE, vol. 86, no. 11, pp. 2278-2324, November 1998.
\bibitem{b24} Y. Sun, X. Huang, D. Kroening, J. Sharp, M. Hill, and R. Ashmore, “Testing deep neural networks,” 2018. arXiv preprint http://arXiv:1803.04792.
\bibitem{b25} R. Banabic, “Techniques for identifying elusive corner-case bugs in systems software,” (No. THESIS). EPFL. 2015. 
\bibitem{b26} T. Ouyang, W. Pedrycz, O. F. Reyes-Galaviz, and N. J. Pizzi, “Granular description of data structures: a two-phase design,” IEEE Transactions on Cybernetics, 2019, Doi: 10.1109/TCYB.2018.2887115
\bibitem{b27} Y. He, A. Kusiak, T. Ouyang, and W. Teng, “Data-driven modeling of truck engine exhaust valve failures: a case study,” Journal of Mechanical Science and Technology, vol. 31, no. 6, 2017, pp. 2747-2757. 
\bibitem{b30} https://github.com/thouyang/ccd-dsa
\bibitem{b28} https://github.com/SemanticApplicationDesignLanguage/sadl
\bibitem{b29} S. Narkhede, “Understanding AUC-ROC Curve,” Towards Data Science, 2018, pp. 26. 


\bibitem{b31} A. Kurakin, I. Goodfellow, and S. Bengio, “Adversarial examples in the physical world,” 2016.  arXiv preprint http://arXiv:1607.02533.
\bibitem{b32} N. Carlini, and D. Wagner, “Towards evaluating the robustness of neural networks,” In 2017 ieee symposium on security and privacy (sp), IEEE, pp. 39-57, May 2017.
\bibitem{b33} A. Madry, A. Makelov, L. Schmidt, D. Tsipras, and A. Vladu, “Towards deep learning models resistant to adversarial attacks,” 2017. arXiv preprint http://arXiv:1706.06083.
\bibitem{b34} D. M. Powers, “Evaluation: from precision, recall and F-measure to ROC, informedness, markedness and correlation,” 2020. arXiv preprint https://arXiv:2010.16061.
\bibitem{b35} F. Riaz, K. Kamal, T. Zafar, and R. Qayyum, “An inspection approach for casting defects detection using image segmentation,” In 2017 International Conference on Mechanical, System and Control Engineering (ICMSC), IEEE, pp. 101-105, May 2017.

\end{thebibliography}
\end{document}